%% file: iclr2027_conference.tex
\documentclass{article} 
\usepackage{iclr2027_conference,times}

\input{math_commands.tex}

\usepackage{hyperref}
\usepackage{url}
\usepackage{xspace}
\usepackage[T1]{fontenc}
\usepackage{enumitem}
\usepackage{tcolorbox}
\usepackage{booktabs}
\usepackage{multirow}
\usepackage{xcolor}
\usepackage{graphicx}
\usepackage{booktabs}
\tcbuselibrary{breakable}
\newcommand{\sys}{\mbox{\textsc{ProVer}}\xspace}
\newcommand{\stitle}[1]{\vspace{0.5em}\noindent{\bf #1}}

\newif\ifcomments
\commentstrue        

\title{Targeting Pivotal Decisions for Credit Assignment in Agentic Reinforcement Learning}

\iclrfinalcopy

\author{%
\textbf{Dongwon  Jung\textsuperscript{\textnormal{1,2}}\thanks{Work done during an internship at Microsoft.} \quad
Hemanth Neelgund Ramesh\textsuperscript{\textnormal{3}} \quad
Yifan Wang\textsuperscript{\textnormal{4}} \quad
Xiaomin Li\textsuperscript{\textnormal{2}}} \\
\textbf{Yuexing Hao\textsuperscript{\textnormal{2}} \quad
Yu Hu\textsuperscript{\textnormal{2}} \quad
Muhao Chen\textsuperscript{\textnormal{1}} \quad
Varun Chandrasekaran\textsuperscript{\textnormal{2}}} \\
\textbf{Andrzej Banburski-Fahey\textsuperscript{\textnormal{2}} \quad
Jaron Lanier\textsuperscript{\textnormal{2}}} \\[0.5em]
\textsuperscript{1}University of California, Davis \enspace
\textsuperscript{2}Microsoft \enspace
\textsuperscript{3}University of Washington \enspace
\textsuperscript{4}Purdue University
}

\iclrfinalcopy 
\begin{document}

\maketitle
\lhead{Preprint}

\begin{abstract}
Group Relative Policy Optimization (GRPO) has become a promising approach for training large language model agents. However, its uniform assignment of trajectory-level advantages to all policy tokens fails to distinguish consequential decisions from less relevant ones, obscuring which intermediate decisions contributed to success. We introduce \sys, a framework that targets potentially pivotal decisions for fine-grained credit assignment in agentic reinforcement learning. Given a rollout group, an agentic judge contrasts successful and failed trajectories to propose a segment potentially responsible for their divergent outcomes. Rather than directly trusting the judge’s assessment, \sys verifies the proposed segment by estimating its advantage from the difference in terminal success rates between current-policy continuations sampled before and after the segment. Positive estimates are then incorporated into the GRPO advantages of policy tokens within the proposed segment. By using model judgment only to select where to verify, \sys grounds local credit in observed outcomes without exhaustively evaluating every intermediate state. 
Across ALFWorld, WebShop, and SearchQA, \sys achieves the strongest average performance at both model scales, with relative improvements over GRPO of 9.91\% and 7.12\% for Qwen3.5-2B and Qwen3.5-4B, respectively. Further analyses demonstrate that informed segment selection improves policy training with modest additional generation overhead, even without a frontier-scale judge model, highlighting the effectiveness and efficiency of selectively targeting pivotal decisions for fine-grained credit assignment in agentic reinforcement learning.
\end{abstract}

\input{sections/intro}

\input{sections/related_work}
\input{sections/method}
\input{sections/experiment_setup}
\input{sections/experiment_result}
\input{sections/conclusion}

\section*{AI use statement}

We used generative AI tools to assist with experiment implementation and debugging, quantitative result analysis, and manuscript revision. In particular, these tools
helped implement and test training and evaluation infrastructure, inspect experimental artifacts, and improve the clarity and structure of the paper. All AI-assisted code was reviewed and tested against the intended behavior, and all reported results and numerical claims were checked against the underlying experiment artifacts or independently recomputed by the authors. We did not use generative AI to create the benchmark
datasets or ground-truth task outcomes. The authors reviewed all AI-assisted work and take responsibility for the final text, claims, code, and artifacts presented in this paper.

\bibliography{iclr2027_conference}
\bibliographystyle{iclr2027_conference}

\input{sections/appendix}

\end{document}

%% file: math_commands.tex
\usepackage{amsmath,amsfonts,bm}
\usepackage{bbm}

\def\eqref#1{equation~\ref{#1}}

\def\1{\bm{1}}

\DeclareMathAlphabet{\mathsfit}{\encodingdefault}{\sfdefault}{m}{sl}
\SetMathAlphabet{\mathsfit}{bold}{\encodingdefault}{\sfdefault}{bx}{n}



%% file: sections/intro.tex
\section{Introduction}

Large language model (LLM) agents have become increasingly competent at solving complex tasks through multi-turn interaction with external tools and environments, including web navigation~\citep{zhou2024webarena,wei2025webagent}, information retrieval~\citep{chen2025learning,jin2025searchr}, and tool use~\citep{huang2024planning,qian2026toolrl,feng2026retool}. Recent work has therefore trained LLM agents directly through environment interaction using reinforcement learning with verifiable outcomes~\citep{wei2025webagent,jin2025searchr,feng2026retool}. In particular, Group Relative Policy Optimization (GRPO) has become a widely used foundation for such training because it estimates relative advantages from multiple trajectories sampled for the same task without requiring a separately trained value model~\citep{shao2024deepseekmath,guo2025deepseek}. However, when applied to long-horizon agents, its trajectory-level supervision creates a fundamental challenge for credit assignment: terminal feedback distinguishes trajectories within a group but provides little information about which intermediate decisions were consequential to the outcome. Thus, the same trajectory-level advantage is assigned to all policy tokens within a trajectory, regardless of their contributions, potentially reinforcing mistakes in successful trajectories and penalizing useful progress in failed ones.

This limitation has motivated approaches that provide finer-grained, intermediate supervision to distinguish the contributions of intermediate decisions rather than uniformly propagating the trajectory-level advantages. Prior work obtains intermediate supervision through learned value functions or critic models that estimate expected returns~\citep{bahdanau2017an,zhou2024archer}, or through process reward models and LLM evaluators that directly assess intermediate behavior~\citep{lightman2024let,zhang2026criticsearch,wang2026enhancing}.
Although these approaches provide finer-grained supervision, their model-based evaluations may be inaccurate, unreliable as the policy distribution shifts, or susceptible to exploitation during optimization~\citep{liu2026save,cui2026process,hou2025treerl}. Monte Carlo policy evaluation instead estimates intermediate-state values by averaging terminal returns from fresh continuations sampled from the current policy, grounding supervision in observed outcomes rather than model-predicted values or evaluator judgments \citep{sutton1988learning,sutton2018reinforcement}. In long-horizon reasoning and agentic settings, this can be implemented by branching from intermediate states and sampling multiple continuations~\citep{kazemnejad2025vineppo,guo2026segment,ji2026tree}. However, extending this approach to long-horizon agentic tasks remains costly since evaluating every candidate state may require repeatedly restoring environment states and executing full continuations. 
This motivates selectively identifying potentially consequential decisions and concentrating fine-grained credit assignment on them, rather than exhaustively evaluating every intermediate state.

To address this challenge, we introduce \sys (\textbf{Pro}pose, \textbf{Ver}ify, Credit), a framework for targeting potentially pivotal decisions for fine-grained credit assignment in agentic reinforcement learning. The key idea is to use model judgment only to identify a candidate segment, then concentrate outcome verification on its boundaries to estimate its advantage from current-policy continuations. \sys comprises three stages: propose, verify, and credit. In the propose stage, an LLM-based agentic judge contrasts successful and failed trajectories from the same GRPO group to localize a potentially pivotal segment in a successful trajectory. 
In the verify stage, we restore the environment states immediately before and after the proposed segment and sample fresh continuations from the current policy. The difference in empirical terminal success rates provides an estimate of the segment's advantage under the current policy. In the credit stage, positive estimates are added to the original GRPO advantages of policy tokens within the proposed segment. By separating LLM-based proposal from Monte Carlo verification, \sys grounds additional credit in observed outcomes rather than potentially inaccurate critic scores, while restricting evaluation to the proposed segment's boundaries.

We evaluate \sys on ALFWorld \citep{shridhar2021alfworld}, WebShop \citep{yao2022webshop}, and SearchQA using Qwen3.5-2B and Qwen3.5-4B.
\sys achieves the strongest average performance at both model scales, yielding relative improvements over GRPO of 9.91\% and 7.12\%, respectively.
Further analyses demonstrate that informed segment selection improves policy learning with modest additional generation overhead, even without a frontier-scale judge model. Together, these results support selective outcome verification as an effective and efficient approach to credit assignment in agentic reinforcement learning.

%% file: sections/related_work.tex
\section{Related Work}
\label{sec:related-work}

\subsection{Agentic Reinforcement Learning}

Agentic reinforcement learning casts interaction with an environment as a sequential decision problem and optimizes policies over complete environment trajectories.
This paradigm spans web navigation, information seeking, and tool use~\citep{wei2025webagent,jin2025searchr,chen2025learning,qian2026toolrl}.
A common foundation is Group Relative Policy Optimization (GRPO), which assigns a trajectory-level outcome advantage to all policy-generated tokens without training a value model~\citep{shao2024deepseekmath,wang2025ragen}.
GiGPO~\citep{feng2026group} refines this supervision through anchor-state grouping, comparing discounted downstream returns of actions taken from repeated environment states.
This provides step-level credit without auxiliary models or additional rollouts, although its coverage depends on state recurrence within the sampled group.
\sys instead estimates local value changes through fresh continuations from judge-selected boundary states.

\subsection{Critic-Based Process Supervision for Credit Assignment}

LLM critics provide intermediate supervision by assessing reasoning steps or agent actions.
Some methods use natural-language critiques to identify errors, guide revised attempts, and incorporate critique-guided improvements into the policy~\citep{zhang2025critique,lin2026icrl}.
Beyond guiding revisions, LLM critics can directly supply intermediate credit.
CriticSearch uses a frozen retrospective critic to evaluate search actions using completed trajectories and reference answers, converting its judgments into turn-level feedback~\citep{zhang2026criticsearch}.
Related methods assign semantic roles to actions and translate them into local credit corrections~\citep{xu2026triage}, or weight retrieval rounds according to their judged contributions~\citep{wang2026enhancing}.
In these credit-assignment methods, evaluator judgments directly shape local credit.
In \sys, the agentic judge selects a candidate segment, while fresh environment continuations determine its estimated advantage and the resulting additional credit.

\subsection{Outcome-Based Credit Assignment}

Outcome-based credit assignment estimates the contribution of intermediate decisions from downstream task outcomes.
Monte Carlo value estimation provides one way to obtain this signal by averaging returns from continuations sampled at intermediate states.
Prior work organizes outcome-labeled continuations as implicit prefix trees or explicitly sampled branching structures~\citep{hou2025treerl,ji2026tree,zhao2026reinforced}.
VinePPO samples continuations at intermediate reasoning-step boundaries to estimate step-level advantages~\citep{kazemnejad2025vineppo}, while SPO partitions trajectories into segments and estimates segment-level advantages through chain- or tree-based sampling~\citep{guo2026segment}.
\sys shares their use of continuation-based value estimation, but uses an agentic judge to select one potentially consequential segment through cross-trajectory comparison, focusing evaluation on its two boundaries rather than evaluating successive steps or segments throughout a trajectory.

%% file: sections/method.tex
\section{Method}
\label{sec:method}

We consider reinforcement learning for LLM agents that receive verifiable terminal rewards from an environment. We first introduce the agent–environment setting and GRPO formulation in Section~\ref{sec:preliminaries}, followed by an overview of \sys in Section~\ref{sec:overview}. We then describe its three stages in Sections~\ref{sec:propose}--\ref{sec:credit}. 

\subsection{Preliminaries}
\label{sec:preliminaries}

\stitle{Agent--environment interaction.}
Let $x$ denote a task instance and $\pi_\theta$ an LLM agent policy.
At turn $t$, the complete agent--environment state $s_t$ comprises the interaction history available to the policy and the corresponding environment state.
The policy samples a response $a_t \sim \pi_\theta(\cdot \mid s_t)$, which induces the next state $s_{t+1}$.
Repeating this interaction produces a trajectory
$\tau = (s_1, a_1, \ldots, s_T, a_T, s_{T+1})$.
After the final turn, the environment returns a terminal reward $R(\tau)$.
We focus on tasks with binary outcomes, where $R(\tau) \in \{0,1\}$ indicates failure or success, and assume no intermediate rewards.

\stitle{Agentic reinforcement learning with GRPO.}
For each task $x$, GRPO samples a group of $G$ trajectories
$\mathcal{G} = \{\tau_i\}_{i=1}^{G}$ from the same policy.
Let $R_i = R(\tau_i)$.
With binary rewards, we use the mean-centered group-relative advantage:
\begin{equation}
    A_i^{\mathrm{GRPO}}
    =
    R_i - \frac{1}{G}\sum_{j=1}^{G} R_j.
    \label{eq:grpo-advantage}
\end{equation}
GRPO assigns $A_i^{\mathrm{GRPO}}$ to every trainable policy token in $\tau_i$, while environment observations and other non-policy tokens are masked from the optimization objective.
This distinguishes successful from failed trajectories within the group, but it does not distinguish among the decisions that compose each trajectory.

We denote the successful and failed trajectories by
$\mathcal{G}^{+} = \{\tau_i \in \mathcal{G} \mid R_i = 1\}$ and
$\mathcal{G}^{-} = \{\tau_i \in \mathcal{G} \mid R_i = 0\}$,
respectively.
We call $\mathcal{G}$ a \emph{mixed-outcome group} when both subsets are nonempty.

\input{resources/main_figure}

\subsection{\sys Overview}
\label{sec:overview}

Figure~\ref{fig:targetcredit_overview} illustrates the overview of \sys. Given a mixed-outcome GRPO group, \sys aims to localize potentially pivotal decisions within a successful trajectory and assign them fine-grained credit without exhaustively evaluating every intermediate state. We operationalize such decisions as short contiguous segments of agent turns and estimate their contribution from downstream outcomes.

\sys proceeds in three stages. In the propose stage, an LLM agent, called the agentic judge, contrasts successful and failed trajectories to select a potentially consequential segment from a successful trajectory. 
In the verify stage, \sys restores the agent–environment states immediately before and after the proposed segment and samples fresh current-policy continuations from both states. The difference between the two empirical terminal success rates estimates the segment’s advantage. In the credit stage, when this estimate is positive, \sys adds it to the original GRPO advantage assigned to each trainable policy token in the selected segment. This design combines model-based guidance~\citep{zhang2026criticsearch,wang2026enhancing} with outcome-grounded Monte Carlo estimation~\citep{kazemnejad2025vineppo,guo2026segment}: the judge identifies a promising segment without directly supplying credit scores, while fresh current-policy continuations estimate its advantage without requiring exhaustive evaluation of intermediate states.

\subsection{Propose: Agentic Judge-Guided Segment Selection}
\label{sec:propose}

The propose stage localizes a potentially consequential decision or sequence of decisions within a successful trajectory. We represent the candidate as a contiguous segment of agent turns, which is subsequently evaluated to determine whether it warrants additional credit. Selecting such a segment requires understanding how its actions address difficulties encountered elsewhere in the rollout group, rather than assessing each action in isolation.
Because the relevant evidence may be scattered across long trajectories, we use an \textbf{agentic judge} that can search for recurring failure patterns and selectively inspect the turns needed to assess a candidate segment.
This allows segment selection to draw on cross-trajectory context without requiring every trajectory to be processed in full.

\stitle{Agentic Judge Design.}
The judge is implemented as an external LLM agent and receives the task, a complete turn-indexed successful trajectory $\tau^+ \in \mathcal{G}^{+}$, and compact previews of the failed trajectories in $\mathcal{G}^{-}$, each truncated to a fixed length.
It uses the failed trajectories to identify recurrent difficulties and examines how a segment in the successful trajectory avoids or resolves them.
To support this inspection, we provide two tools adapted from \cite{lee2026agentic}:
\begin{itemize}
[leftmargin=*,noitemsep,nolistsep,itemsep=3pt]
    \item \textbf{\texttt{search\_trajectory(query, k)}}: Searches all failed trajectories using the natural-language \texttt{query} and returns the top-$k$ bounded turn previews with their trajectory identifiers and turn indices.
    \item \textbf{\texttt{get\_segment(traj\_id, start\_turn, end\_turn)}}: Retrieves the inclusive turn range $[\texttt{start\_turn}, \texttt{end\_turn}]$ from the trajectory identified by \texttt{traj\_id}, including the policy-visible context before the first retrieved turn.
\end{itemize}
The judge can alternate between searching for supporting evidence and inspecting relevant turns to refine its hypothesis about which segment warrants evaluation.
After inspection, it outputs inclusive turn boundaries $(\ell,r)$ in $\tau^+$, defining the proposed segment
$\tau^{\mathrm{seg}}=(s_{\ell},a_{\ell},\ldots,s_r,a_r,s_{r+1})$. The selected segment therefore remains a hypothesis about which decisions contributed to success; the verify stage tests whether it warrants additional credit.

\subsection{Verify: Outcome-Based Segment Evaluation}
\label{sec:verify}

The propose stage identifies a potentially consequential segment, but it does not determine whether the segment deserves additional credit.
The verify stage makes this distinction explicit: the judge's proposal only allocates the evaluation budget, while the segment's advantage is estimated from downstream task outcomes.
Following the Monte Carlo principle of estimating intermediate values from sampled continuations, we compare fresh current-policy continuations from the states immediately before and after the proposed segment.
The difference between the two empirical terminal success rates provides an outcome-based estimate of whether, and by how much, executing the segment improves the policy's probability of success.
We first formalize the segment advantage that this comparison seeks to estimate.

\stitle{Segment advantage.}
The turn boundaries $(\ell,r)$ determine the pre- and post-segment states $s_{\mathrm{pre}}=s_\ell$ and $s_{\mathrm{post}}=s_{r+1}$.
Let $\mathbf{a}_{\ell:r}=(a_\ell,\ldots,a_r)$ denote the recorded action sequence within $\tau^{\mathrm{seg}}$.
Let $\pi$ denote the policy that generated the source group, held fixed during verification.
We write $V^\pi(s)$ for the expected discounted return when following $\pi$ from state $s$, and $Q_{\mathrm{seg}}^\pi(s_{\mathrm{pre}},\mathbf{a}_{\ell:r})$ for the expected discounted return when executing the recorded segment and then following $\pi$.

Treating the recorded segment as a temporally extended action and assuming deterministic segment execution, its current-policy advantage is given below.
The final equality uses our setting of nonterminal segments with no intermediate rewards and $\gamma=1$:
\begin{equation}
\begin{aligned}
    A_{\mathrm{seg}}^\pi(s_{\mathrm{pre}},\mathbf{a}_{\ell:r})
    &= Q_{\mathrm{seg}}^\pi(s_{\mathrm{pre}},\mathbf{a}_{\ell:r})
       - V^\pi(s_{\mathrm{pre}}) \\
    &= \sum_{j=\ell}^{r}\gamma^{j-\ell}r_j
       + \gamma^d V^\pi(s_{\mathrm{post}})
       - V^\pi(s_{\mathrm{pre}}) \\
    &= V^\pi(s_{\mathrm{post}})-V^\pi(s_{\mathrm{pre}}),
\end{aligned}
    \label{eq:segment-value-difference}
\end{equation}
where $r_j$ is the reward received at turn $j$, $d=r-\ell+1$ is the segment duration in turns, and $\gamma$ is the per-turn discount factor.
With binary terminal rewards and $\gamma=1$, the boundary values represent probabilities of eventual success.
Thus, verifying a proposal reduces to estimating the current-policy values at these two boundaries.

\stitle{Boundary rollouts.}
We estimate the two boundary values through Monte Carlo policy evaluation.
For each valid proposal, we restore $s_{\mathrm{pre}}$ and $s_{\mathrm{post}}$, including the interaction history and corresponding environment state.
From each restored state, we independently sample $K$ fresh continuations using the fixed policy $\pi$.
Let $R_k^{\mathrm{pre}}$ and $R_k^{\mathrm{post}}$ denote their terminal rewards.
Because these rewards are binary, their empirical means estimate the policy's success probabilities at the two boundaries:
\begin{equation}
    \widehat{V}_{\mathrm{pre}}^{\pi}
    = \frac{1}{K}\sum_{k=1}^{K}R_k^{\mathrm{pre}},
    \qquad
    \widehat{V}_{\mathrm{post}}^{\pi}
    = \frac{1}{K}\sum_{k=1}^{K}R_k^{\mathrm{post}},
    \qquad
    \widehat{\Delta}_{\mathrm{seg}}
    = \widehat{V}_{\mathrm{post}}^{\pi}
    - \widehat{V}_{\mathrm{pre}}^{\pi},
    \label{eq:boundary-values}
\end{equation}
where $\widehat{\Delta}_{\mathrm{seg}}$ is the resulting segment-advantage estimate.

Under the stated assumptions, $\widehat{\Delta}_{\mathrm{seg}}$ is conditionally unbiased for the segment advantage in Equation~\ref{eq:segment-value-difference}, given the selected segment and its boundary states.\footnote{Appendix~\ref{app:segment-estimator} proves this result and clarifies its scope.}
A positive $\widehat{\Delta}_{\mathrm{seg}}$ provides outcome-based evidence that the segment improves the policy's probability of success.

\subsection{Credit: Segment-Level Advantage Assignment}
\label{sec:credit}

Because the boundary-value comparison estimates the advantage of the segment as a whole, we assign the same verified bonus to all of its trainable policy tokens.
Let $u$ index trainable policy tokens in $\tau^+$, and let
$I_u^{\mathrm{seg}}=\mathbbm{1}[\ell\leq\operatorname{turn}(u)\leq r]$
indicate whether token $u$ belongs to an action turn within the proposed segment.
For a valid segment-advantage estimate, we define
\begin{equation}
    A_u^{\scriptstyle\textsc{\sys}}
    =
    \begin{cases}
        A^{\mathrm{GRPO}}
        + \lambda\widehat{\Delta}_{\mathrm{seg}}I_u^{\mathrm{seg}},
        & \text{if }\widehat{\Delta}_{\mathrm{seg}}>0, \\
        A^{\mathrm{GRPO}}, & \text{otherwise},
    \end{cases}
    \label{eq:targetcredit-advantage}
\end{equation}
where $A^{\mathrm{GRPO}}$ is the trajectory-level advantage of $\tau^+$ and $\lambda\geq0$ controls the strength of the segment-level bonus.
Invalid proposals or measurements receive no additional credit.
We substitute $A_u^{\scriptstyle\textsc{\sys}}$ for the ordinary GRPO advantage on $\tau^+$ while leaving the source trajectories, policy ratios, clipping rule, and other optimization terms unchanged.
All other trajectories retain their original GRPO advantages.
Boundary continuations are used only for advantage estimation and are not included in the policy-training batch. 

Overall, the judge selects potentially pivotal segments for evaluation, while boundary-rollout outcomes determine whether and how much additional credit they receive.
This separation enables \sys to concentrate outcome-based evaluation on a small part of the trajectory and augment GRPO’s trajectory-level supervision with selectively verified local credit.

%% file: resources/main_figure.tex
\begin{figure*}[t]
    \centering
    \includegraphics[width=\textwidth]{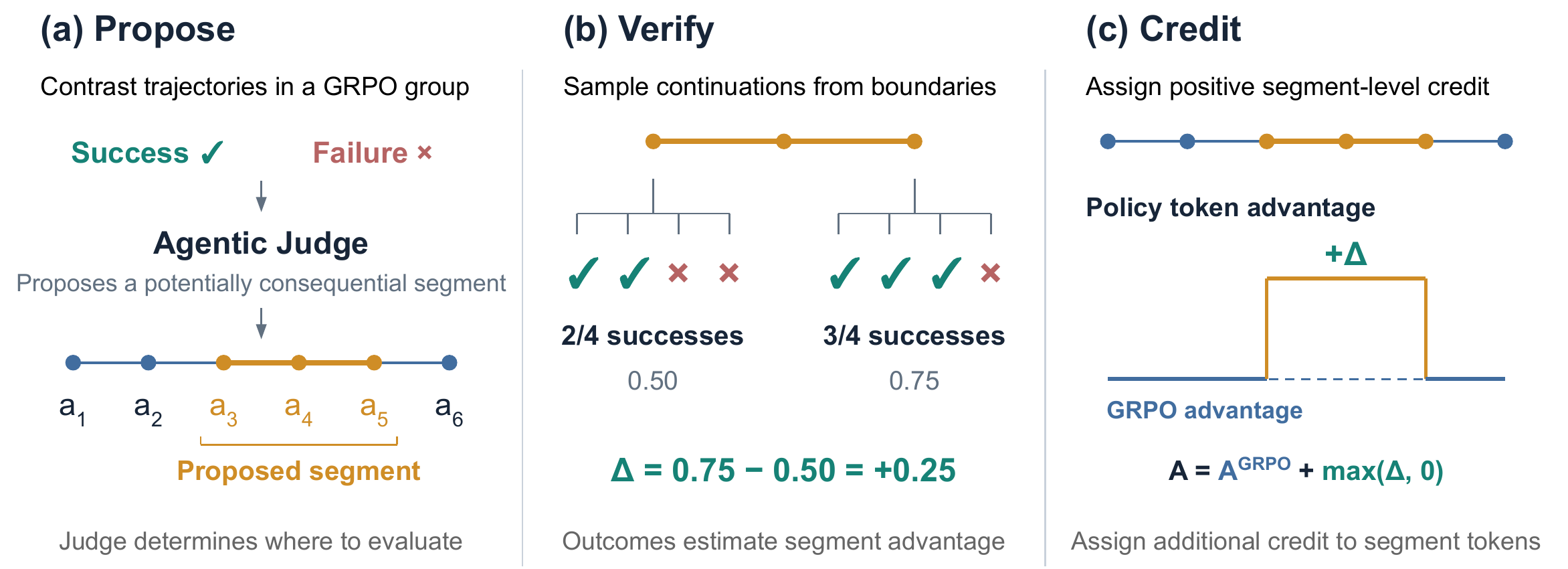}
\caption{
Overview of \textsc{\sys}. \textbf{(a) Propose:} An agentic judge contrasts successful and failed trajectories to localize a potentially consequential segment. \textbf{(b) Verify:} Boundary continuations estimate the segment advantage $\Delta$ from the difference in terminal success rates.
\textbf{(c) Credit:} Positive estimates augment GRPO advantages only for policy tokens within the selected segment. 
}
\label{fig:targetcredit_overview}
\vspace{-0.6em}
\end{figure*}

%% file: sections/experiment_setup.tex
\section{Experimental Setup}
\label{sec:experimental-setup}

\subsection{Agent Environments}
We evaluate \sys on three long-horizon agent environments: (1) \textbf{ALFWorld}~\citep{shridhar2021alfworld}: a text-based household simulator in which the agent observes its surroundings and executes natural-language actions to manipulate objects and complete a task; (2) \textbf{WebShop}~\citep{yao2022webshop}: a simulated e-commerce website in which the agent issues searches and navigates product pages through web actions before selecting a product that satisfies a natural-language goal; and (3) \textbf{SearchQA}: a corpus-backed retrieval environment in which the agent iteratively submits search queries, examines retrieved evidence, and returns an answer to a single- or multi-hop question. Further details on the environments, interaction protocols, and prompt templates are provided in Appendix~\ref{app:agent-environments}.

\subsection{Baselines}

We compare \sys with six baselines:
(1) \textbf{GRPO}~\citep{shao2024deepseekmath}, the standard group-relative policy optimization baseline;
(2) \textbf{Budget-Matched GRPO}, which augments GRPO with additional policy rollouts to match the rollout budget used by \sys for segment-boundary value estimation;
(3) \textbf{GiGPO}~\citep{feng2026group}, which augments GRPO with step-level advantages derived from discounted returns at repeated environment states;
(4) \textbf{SPO-chain}~\citep{guo2026segment}, which assigns segment-level credit using differences between Monte Carlo value estimates at segment boundaries;
(5) \textbf{SPO-tree}~\citep{guo2026segment}, which uses tree-structured sampling to share trajectory prefixes and estimate sibling-relative advantages;
(6) \textbf{CriticSearch}~\citep{zhang2026criticsearch}, which augments GRPO with binary action-level scores from a retrospective LLM critic.
We use the same underlying LLM for this critic and our agentic judge.

These baselines examine complementary aspects of credit assignment.
Budget-Matched GRPO tests whether additional policy rollouts alone account for the gains.
GiGPO provides a comparison with step-level credit inferred from repeatedly visited states.
SPO-chain and SPO-tree compare targeted segment selection with fixed segment boundaries and tree-structured sampling, respectively.
CriticSearch compares credit derived from sampled downstream outcomes with credit derived directly from an LLM critic's action-level judgments.
Complete training and method-specific implementation details are provided in Appendix~\ref{app:baseline-details}.

\subsection{Implementation Details}

We use Prime-RL~\citep{primeintellect2025prime-rl} to train Qwen3.5-2B and Qwen3.5-4B for 100 optimizer steps.
The standard GRPO configuration uses 16 groups of $G=8$ trajectories per update.
By default, \sys uses \texttt{gpt-5.4-mini} as its agentic judge, matching the underlying LLM used by the CriticSearch critic.
We apply \sys to groups with success rates strictly between 0 and $0.5$, selecting one successful trajectory and proposing one nonterminal segment of at most four action turns per eligible group.
Verification uses $K=8$ continuations per boundary, and positive segment advantages are incorporated with weight $\lambda=1$.
We evaluate task success rate on ALFWorld and WebShop and semantic answer accuracy on SearchQA, using \texttt{gpt-5.4-mini} to judge answer correctness.
Additional training and segment-selection details are provided in Appendix~\ref{app:implementation-details}.

%% file: sections/experiment_result.tex
\section{Experimental Results}
\label{sec:experimental-results}

\input{resources/main_result_table}

\subsection{Main Results}

\stitle{\sys delivers the strongest overall performance.}
Table~\ref{tab:main-results} shows that \sys achieves the highest average score at both model scales and the best result in four of the six model--environment settings.
Relative to GRPO, \sys improves average performance by 9.91\% with Qwen3.5-2B and 7.12\% with Qwen3.5-4B, with gains in all six settings.
These results demonstrate the effectiveness of augmenting trajectory-level GRPO with fine-grained credit targeted toward potentially consequential decisions.

\stitle{Additional group-level rollouts alone do not explain the gains.}
Budget-Matched GRPO matches the number of additional policy rollouts used by \sys but retains trajectory-level credit.
\sys outperforms this control in every setting, with relative improvements ranging from 5.34\% to 22.23\%.
Moreover, Budget-Matched GRPO does not consistently outperform standard GRPO.
These results support concentrating the additional rollouts on evaluating potentially consequential segments, rather than simply collecting more trajectories for GRPO.

\stitle{Targeting consequential decisions outperforms structured rollout baselines.}
\sys outperforms SPO-tree across all six settings and SPO-chain in five of the six settings.
The gains over SPO-chain are particularly pronounced on SearchQA, reaching 43.95\% with Qwen3.5-2B and 17.74\% with Qwen3.5-4B.
While SPO-chain evaluates fixed segment boundaries and SPO-tree uses a predetermined branching structure, \sys evaluates the boundaries of a segment selected by the agentic judge.
These comparisons support the effectiveness of using judge-guided localization to concentrate continuation-based credit estimation on potentially consequential decisions.

\stitle{Judge-guided verification outperforms direct critic-based credit.}
CriticSearch assigns action-level credit directly from binary labels produced by a retrospective LLM critic, whereas \sys uses an agentic judge to select a segment and estimates its advantage from downstream outcomes.
Both use the same underlying LLM for the critic or judge.
\sys achieves a higher average score at both model scales and outperforms CriticSearch in five of the six settings.
These results support the proposed separation between segment selection and outcome-based credit estimation, compared with directly translating critic judgments into credit.

\subsection{Training Cost Analysis}

Having established the performance gains of \sys, we next examine the additional computation required to obtain them.
Table~\ref{tab:cost-analysis} reports generated policy tokens, judge API cost, and optimizer-step wall time for Qwen3.5-4B training.

\stitle{Selective evaluation of potentially consequential decisions limits additional policy generation.}
\sys selects one short segment from one successful trajectory per eligible group and evaluates only its two boundary states.
SPO-chain instead evaluates multiple boundaries for every eligible successful trajectory, while CriticSearch scores every scorable turn in all eight trajectories.
Relative to GRPO, \sys increases generated policy tokens per step by 2.4\% on ALFWorld, 11.6\% on WebShop, and 16.8\% on SearchQA.
Together with the strongest average performance in Table~\ref{tab:main-results}, these results show that selectively evaluating one potentially consequential segment provides fine-grained credit with modest additional policy-generation overhead.

\stitle{\sys achieves the lowest per-step wall time among the evaluated fine-grained methods.}
Table~\ref{tab:cost-analysis} shows that \sys has lower per-step wall time than SPO-tree, SPO-chain, and CriticSearch across all three environments.
This advantage is not explained by generation volume alone: SPO-tree generates fewer policy tokens than GRPO on ALFWorld and WebShop but takes substantially longer per optimizer step, consistent with additional execution overhead from sequential branching and environment-state restoration.
SPO-chain incurs additional rollout cost by evaluating multiple segment boundaries in every eligible successful trajectory, whereas \sys evaluates only two boundaries of one selected segment.
Similarly, CriticSearch scores every scorable action across all eight trajectories in an eligible group, whereas the agentic judge in \sys selectively inspects trajectory evidence to propose one segment, with 38.4\% lower judge API cost on average across the three environments.
Together, these comparisons show that concentrating evaluation on a single proposed segment can provide segment-level credit with lower per-step wall time than the evaluated fine-grained alternatives.

\input{resources/cost_analysis_table}

\subsection{Effect of Judge Model Choice}
\label{sec:agentic-judge-ablation}

A natural question is whether \sys requires an expensive frontier-scale model as its external judge.
We compare three models from the GPT-5.4 family and a locally served Qwen3.5-9B judge while holding the Qwen3.5-4B policy backbone, training procedure, and outcome-verification mechanism fixed.
We also include a \texttt{Random} baseline that replaces the judge with random sampling of a valid nonterminal segment from the same selected successful trajectory, while retaining the verification budget and credit-assignment rule.\footnote{The random sampling procedure is detailed in Appendix~\ref{app:baseline-details}.}
Table~\ref{tab:agentic-judge-ablation} reports policy performance across the three environments, together with proposal acceptance rate ($\hat{\Delta}_{\mathrm{seg}}>0$) and mean segment length on SearchQA.

\stitle{Informed segment selection concentrates evaluation on more promising candidate segments.}
All four agentic judges outperform random selection across all three environments.
On SearchQA, random selection yields an acceptance rate of 19.5\%, compared with 58.5--72.4\% for the agentic judges, despite proposing substantially longer segments on average (2.38 vs.\ 1.03--1.45 turns).
This suggests that the judges more precisely localize potentially consequential segments, rather than obtaining positive estimates simply by covering more actions.
Together with the policy results, this supports the benefit of judge-guided localization over allocating fine-grained credit to arbitrarily selected segments.

Within the GPT-5.4 family, the lower acceptance rates of \texttt{gpt-5.4-mini} and \texttt{gpt-5.4} accompany shorter proposals than those of \texttt{gpt-5.4-nano}.
A plausible explanation is that these models follow the minimal-segment instruction more closely, which requires more precise localization, whereas longer segments span more environment interactions and observations and may be more likely to yield a positive $\hat{\Delta}_{\mathrm{seg}}$.
However, lower acceptance does not necessarily imply less effective selection, as \texttt{gpt-5.4-mini} produces stronger downstream policies than \texttt{gpt-5.4-nano} across all three environments.

\stitle{Effective segment proposal does not require a frontier-scale judge.}
Although informed selection matters, using a frontier-scale model does not necessarily improve the resulting policy.
Both \texttt{gpt-5.4-mini} and \texttt{gpt-5.4-nano} improve over standard GRPO across all three environments, and \texttt{gpt-5.4-mini} achieves the highest scores on WebShop and SearchQA.
The locally served Qwen3.5-9B judge achieves the highest reported ALFWorld score and exceeds GRPO on SearchQA, although it falls below GRPO on WebShop.
In contrast, \texttt{gpt-5.4} does not achieve the highest score on any environment.
Together, these findings support the practical design of \sys: smaller or locally served models can identify useful candidate segments, while environment rollouts supply the numerical credit signal.
Judge choice remains task-dependent, but frontier-scale capability is unnecessary for strong performance in the evaluated settings.

\input{resources/agentic_judge_ablation_table}

%% file: resources/main_result_table.tex
\begin{table}[t]
    \centering
    \begin{tabular}{lcccc}
        \toprule
        Method & ALFWorld & WebShop & SearchQA & Avg. \\
        \midrule
        \multicolumn{5}{l}{\textit{Qwen3.5-2B}} \\
        \midrule
        GRPO
        & 84.08 {\scriptsize $\pm$ 3.02}
        & 42.53 {\scriptsize $\pm$ 0.31}
        & 35.58 {\scriptsize $\pm$ 0.38}
        & 54.07 {\scriptsize $\pm$ 0.78} \\
        Budget-Matched GRPO
        & 84.05 {\scriptsize $\pm$ 4.13}
        & 39.00 {\scriptsize $\pm$ 2.62}
        & 37.75 {\scriptsize $\pm$ 0.25}
        & 53.60 {\scriptsize $\pm$ 1.32} \\
        GiGPO
        & 85.82 {\scriptsize $\pm$ 2.24}
        & 44.60 {\scriptsize $\pm$ 0.35}
        & 36.17 {\scriptsize $\pm$ 1.66}
        & 55.53 {\scriptsize $\pm$ 1.25} \\
        SPO-tree
        & 56.96 {\scriptsize $\pm$ 1.14}
        & 38.47 {\scriptsize $\pm$ 0.50}
        & 31.67 {\scriptsize $\pm$ 0.72}
        & 42.36 {\scriptsize $\pm$ 0.27} \\
        SPO-chain
        & 75.87 {\scriptsize $\pm$ 1.72}
        & \textbf{47.80} {\scriptsize $\pm$ 2.62}
        & 27.67 {\scriptsize $\pm$ 0.76}
        & 50.45 {\scriptsize $\pm$ 1.20} \\
        CriticSearch
        & 85.08 {\scriptsize $\pm$ 1.49}
        & 39.20 {\scriptsize $\pm$ 0.72}
        & 38.50 {\scriptsize $\pm$ 0.90}
        & 54.26 {\scriptsize $\pm$ 0.63} \\
        \textbf{\sys}
        & \textbf{90.80} {\scriptsize $\pm$ 3.02}
        & 47.67 {\scriptsize $\pm$ 0.42}
        & \textbf{39.83} {\scriptsize $\pm$ 2.08}
        & \textbf{59.43} {\scriptsize $\pm$ 0.74} \\
        \midrule
        \multicolumn{5}{l}{\textit{Qwen3.5-4B}} \\
        \midrule
        GRPO
        & 93.78 {\scriptsize $\pm$ 1.55}
        & 45.67 {\scriptsize $\pm$ 0.58}
        & 40.92 {\scriptsize $\pm$ 0.58}
        & 60.12 {\scriptsize $\pm$ 0.50} \\
        Budget-Matched GRPO
        & 93.28 {\scriptsize $\pm$ 1.29}
        & 44.93 {\scriptsize $\pm$ 0.42}
        & 41.17 {\scriptsize $\pm$ 1.38}
        & 59.79 {\scriptsize $\pm$ 0.94} \\
        GiGPO
        & 97.01 {\scriptsize $\pm$ 0.00}
        & 40.87 {\scriptsize $\pm$ 0.81}
        & 41.83 {\scriptsize $\pm$ 0.95}
        & 59.90 {\scriptsize $\pm$ 0.57} \\
        SPO-tree
        & 92.30 {\scriptsize $\pm$ 3.12}
        & 46.00 {\scriptsize $\pm$ 1.31}
        & 42.92 {\scriptsize $\pm$ 0.72}
        & 60.40 {\scriptsize $\pm$ 0.73} \\
        SPO-chain
        & 94.28 {\scriptsize $\pm$ 1.55}
        & 47.47 {\scriptsize $\pm$ 0.92}
        & 39.92 {\scriptsize $\pm$ 0.80}
        & 60.55 {\scriptsize $\pm$ 0.71} \\
        CriticSearch
        & 95.27 {\scriptsize $\pm$ 0.43}
        & \textbf{48.07} {\scriptsize $\pm$ 1.14}
        & 43.50 {\scriptsize $\pm$ 1.00}
        & 62.28 {\scriptsize $\pm$ 0.35} \\
        \textbf{\sys}
        & \textbf{98.26} {\scriptsize $\pm$ 1.88}
        & 47.93 {\scriptsize $\pm$ 0.61}
        & \textbf{47.00} {\scriptsize $\pm$ 0.90}
        & \textbf{64.40} {\scriptsize $\pm$ 0.92} \\
        \bottomrule
    \end{tabular}
    \caption{Performance of the baselines and \sys with Qwen3.5-2B and Qwen3.5-4B on three benchmarks. We report the mean and sample standard deviation over three evaluation runs.}
    \label{tab:main-results}
    \vspace{-1em}
\end{table}

%% file: resources/cost_analysis_table.tex
\begin{table*}[t]
\centering
\small
\setlength{\tabcolsep}{3pt}
\begin{tabular}{@{}lccccccccc@{}}
\toprule
\multirow{2}{*}{Method}
& \multicolumn{3}{c}{Generated Tokens / Step (K)}
& \multicolumn{3}{c}{Judge Cost / Step (\textcent)}
& \multicolumn{3}{c}{Time / Step (min)} \\
\cmidrule(lr){2-4}\cmidrule(lr){5-7}\cmidrule(l){8-10}
& ALF
& WS
& SQA
& ALF
& WS
& SQA
& ALF
& WS
& SQA \\
\midrule
GRPO
& 237.0
& 97.1
& 37.6
& --
& --
& --
& 3.39
& 0.82
& 1.39 \\
\midrule
SPO-tree
& \textbf{153.2}
& \textbf{71.0}
& 63.4
& --
& --
& --
& 7.18
& 4.85
& 4.65 \\
SPO-chain
& 421.2
& 172.3
& 72.3
& --
& --
& --
& 6.73
& 5.07
& 4.77 \\
CriticSearch
& 227.1
& 136.3
& 45.8
& 17.31
& \textbf{8.54}
& 7.40
& 4.64
& 3.15
& 3.25 \\
\sys
& 242.6
& 108.4
& \textbf{43.9}
& \textbf{4.00}
& 10.81
& \textbf{5.69}
& \textbf{3.73}
& \textbf{2.00}
& \textbf{2.25} \\
\bottomrule
\end{tabular}
\caption{Per-step training cost for Qwen3.5-4B on ALFWorld (ALF), WebShop
(WS), and SearchQA (SQA). Generated tokens count policy-rollout outputs.
Judge cost is calculated from recorded \texttt{gpt-5.4-mini} API usage using
OpenAI's published pricing as of August 2026. Time is the mean wall-clock time per optimizer step in minutes.}
\label{tab:cost-analysis}
\vspace{-1.3em}
\end{table*}

%% file: resources/agentic_judge_ablation_table.tex
\begin{table*}[t]
\centering
\small
\setlength{\tabcolsep}{5pt}
\begin{tabular}{@{}lccccc@{}}
\toprule
\multirow{2}{*}{Agentic Judge}
& \multicolumn{3}{c}{Policy Performance (\%)}
& \multicolumn{2}{c}{SearchQA Proposal Diagnostics} \\
\cmidrule(lr){2-4}\cmidrule(l){5-6}
& ALFWorld
& WebShop
& SearchQA
& Accept (\%)
& Mean Len. \\
\midrule
\texttt{Random}
& 94.0
& 42.4
& 42.8
& 19.5
& 2.38 \\
\texttt{Qwen3.5-9B}
& \textbf{98.5}
& 44.2
& 45.8
& 71.2
& 1.45 \\
\texttt{gpt-5.4-nano}
& 94.8
& 46.2
& 43.5
& 72.4
& 1.25 \\
\texttt{gpt-5.4-mini}
& 98.3
& \textbf{47.9}
& \textbf{47.0}
& 62.5
& 1.18 \\
\texttt{gpt-5.4}
& 96.3
& 47.2
& 43.8
& 58.5
& 1.03 \\
\bottomrule
\end{tabular}
\caption{
Effect of segment selection and judge model choice on Qwen3.5-4B policy training.
\texttt{Random} replaces judge-guided selection with random segment sampling while retaining the same outcome-verification and credit-assignment procedure.
We additionally report proposal diagnostics on SearchQA: acceptance rate ($\widehat{\Delta}_{\mathrm{seg}}>0$) and mean segment length in agent turns.
}
\label{tab:agentic-judge-ablation}
\vspace{-1em}
\end{table*}

%% file: sections/conclusion.tex


\section{Conclusion}
We present \sys, a framework for targeting potentially pivotal decisions for fine-grained credit assignment in agentic reinforcement learning. An agentic judge localizes a candidate segment, while outcome verification estimates its advantage from fresh current-policy continuations.
Incorporating positive segment-advantage estimates into GRPO yields the strongest average performance among the evaluated baselines at both model scales across three agent environments, with modest additional generation overhead.
Our analyses show that informed segment selection contributes to these gains without requiring a frontier-scale judge.
Together, these findings support a practical role for LLM judgment in agentic reinforcement learning: identifying where fine-grained credit is most useful, while leaving the numerical credit signal to downstream environment outcomes.

%% file: sections/appendix.tex
\appendix

\input{sections/appendix/experiment_details}

\input{sections/appendix/conditional_unbiasedness}

%% file: sections/appendix/experiment_details.tex
\section{Experiment Details}
\subsection{Agent Environment Details}
\label{app:agent-environments}

We use the AgentGym implementations of ALFWorld, WebShop, and SearchQA \citep{xi2025agentgym}. The system and user prompts below are reproduced verbatim from our implementation; brace-delimited fields indicate dynamic content in the environment-observation templates. The policy responds through native tool calling with exactly one \texttt{act(action=...)} call per turn.

\paragraph{ALFWorld.}
ALFWorld~\citep{shridhar2021alfworld} is a text-based embodied environment for household tasks. The initial observation contains the task description, current location, visible objects, and exact available actions. Each subsequent observation reports the result of the previous action and an updated action set. The agent must choose an exact available action string, which may navigate to a location or manipulate an object through operations such as taking, opening, placing, heating, cooling, or cleaning. We use the text-world version with a maximum of 50 turns. We train on the canonical 2,620 training games and evaluate on all 134 valid-unseen games. The evaluation metric is task success accuracy.

\input{resources/alfworld_prompt}

\paragraph{WebShop.}
WebShop~\citep{yao2022webshop} presents a natural-language shopping instruction and a rendered e-commerce page. Each observation contains the visible page text and the currently available interaction affordances. The agent may issue \texttt{search[query]} when a search bar is available or select an exact visible target with \texttt{click[target]}. Through these actions, it navigates search results and product pages, chooses product options, and completes a purchase. We allow at most 15 turns. We train on the 10,587 canonical training goals using binary task success as the reward. We evaluate on all 500 canonical test goals and report exact binary task success metrics. 

\input{resources/webshop_prompt}

\paragraph{SearchQA.}
SearchQA presents a question and allows the agent to alternate between evidence retrieval and answer submission. A \texttt{<search>query</search>} action returns retrieved passages inside the next observation, which the agent can use to formulate later queries. A terminal \texttt{<answer>answer</answer>} action submits a concise final answer. We utilize a FAISS index \citep{douze2024faiss} over the 2018 Wikipedia corpus with E5-base-v2 query embeddings used by \cite{jin2025searchr}. Agents are allowed at most 30 turns to complete a task.

The training pool contains 169,615 questions from the training splits of \citep{jin2025searchr} consisting of NQ \citep{kwiatkowski2019natural} and HotpotQA \citep{yang2018hotpotqa}. Evaluation uses a fixed stratified manifest of 400 held-out questions used by \cite{lin2026icrl}: 50 each from NQ, TriviaQA \citep{joshi2017triviaqa}, PopQA \citep{mallen2023not}, HotpotQA, 2WikiMultiHopQA \citep{ho2020constructing}, and Bamboogle \citep{press2023measuring}, and 100 from MuSiQue \citep{trivedi2022musique}. Training uses a binary normalized exact-match reward. For evaluation, we report semantic-equivalence accuracy using GPT-5.4-mini as an LLM judge to recognize correct answer variants that exact matching may miss.

\input{resources/searchqa_prompt}

\subsection{Additional Implementation Details}
\label{app:implementation-details}

\stitle{Training configuration.}
The standard GRPO configuration has a batch size of 128 trajectories, organized into 16 groups of eight trajectories.
We use AdamW with a learning rate of $10^{-6}$ and weight decay of 0 on ALFWorld and WebShop and 0.1 on SearchQA.
Training uses a sampling temperature of 1 and a maximum completion length per turn of 512 tokens on ALFWorld and WebShop and 256 tokens on SearchQA.
Each run uses four A100 SXM GPUs for policy inference and four for training.
The baselines use the same policy backbones and shared optimization settings, with method-specific changes to credit assignment and rollout allocation detailed in Appendix~\ref{app:baseline-details}.

\stitle{Segment selection and verification.}
For $G=8$, the eligibility criterion selects groups containing one, two, or three successful trajectories.
From each eligible group, we select the successful trajectory with the shortest horizon and ask the judge to identify a contiguous nonterminal segment of at most four action turns.
We restore the exact agent--environment states immediately before and after the segment and sample $K=8$ current-policy continuations from each boundary.
The difference between their mean terminal rewards estimates the segment advantage.
When this estimate is positive, we add it to the GRPO advantage of each trainable policy token within the segment, using $\lambda=1$ in Equation~\ref{eq:targetcredit-advantage}.
All other source tokens retain their GRPO advantages, and ineligible groups retain standard GRPO updates.
Boundary continuations are used only for estimation and are not included in the policy-training batch.

\subsection{Baseline Implementation}
\label{app:baseline-details}




\paragraph{GRPO.}
GRPO assigns Equation~\ref{eq:grpo-advantage} uniformly to all trainable assistant tokens in a trajectory. It uses neither localized credit nor auxiliary policy rollouts and therefore provides the outcome-only reference.

\paragraph{Budget-Matched GRPO.}
Budget-Matched GRPO uses the same outcome-only credit in Equation~\ref{eq:grpo-advantage} but replaces \sys's boundary continuations with additional rollouts. Each update contains 16 groups, and we use two phase-specific group sizes to match the corresponding \sys run's realized rollout budget over 100 steps. Table~\ref{tab:budget-matched-rollouts} reports the number of trajectories sampled per group in each phase. Thus, Budget-Matched GRPO controls for additional policy sampling without using boundary restoration or localized credit.

\begin{table}[h]
	\centering
	\small
	\begin{tabular}{llcccc}
		\toprule
		Model & Benchmark & Phase 1 steps & Rollouts/group & Phase 2 steps & Rollouts/group \\
		\midrule
		Qwen3.5-2B & ALFWorld & 1--84 & 11 & 85--100 & 12 \\
		Qwen3.5-2B & WebShop  & 1--94 & 10 & 95--100 & 11 \\
		Qwen3.5-2B & SearchQA & 1--70 & 10 & 71--100 & 11 \\
		Qwen3.5-4B & ALFWorld & 1--26 & 8  & 27--100 & 9 \\
		Qwen3.5-4B & WebShop  & 1--35 & 10 & 36--100 & 11 \\
		Qwen3.5-4B & SearchQA & 1--36 & 9  & 37--100 & 10 \\
		\bottomrule
	\end{tabular}
	\caption{Number of trajectories sampled per group in each phase of Budget-Matched GRPO. Every update contains 16 groups.}
	\label{tab:budget-matched-rollouts}
\end{table}

\paragraph{GiGPO.}
GiGPO augments trajectory-level GRPO advantages with step-level advantages obtained by grouping action turns that share the same environment state and comparing their discounted downstream returns.
This provides local credit from state recurrence within the sampled rollout group, without auxiliary continuation rollouts or an external judge.
We use the same policy backbones and shared training configuration as the other baselines.

\paragraph{SPO-chain.}
SPO-chain assigns segment $j$ the advantage $A_j=\gamma V_j-V_{j-1}$, where $V_{j-1}$ and $V_j$ are Monte Carlo values at the segment's start and end boundaries and $\gamma$ is the discount factor. It masks tokens with generating-policy probability at least $0.9$. Applying the original method to every trajectory in every group is impractical for long, stateful agent trajectories. We therefore apply it only to successful trajectories in groups containing one, two, or three successes following \sys's setup and divide their action turns into up to three contiguous, approximately equal segments; all other trajectories retain GRPO credit. Let $V_0$ be the original group's mean reward, $V_1$ and $V_2$ the internal-boundary values, and $V_3=R_i$ the trajectory reward. With no temporal discounting ($\gamma=1$), the three advantages are $A_1=V_1-V_0$, $A_2=V_2-V_1$, and $A_3=R_i-V_2$.
Each internal value uses $K=8$ exact-state continuations. Fixed thirds keep the budget comparable to \sys, although SPO-chain remains more expensive because it measures every successful trajectory rather than one selected success. 

\paragraph{SPO-tree.}
SPO-tree constructs a three-level binary tree with branching factors $2$--$2$--$2$, giving 14 sampled branches and eight terminal leaves per training group. For each child $c$ of parent $p$, it assigns the branch advantage $A(c)=V(c)-V(p)$ with no temporal discounting, where a leaf value is its binary terminal reward and each internal value is the mean reward of its descendant leaves. The first two levels advance by four environment actions per level in ALFWorld and one action per level in WebShop and SearchQA, while the final level continues to the terminal condition or episode horizon. Each child resumes from the parent's exact environment state. We pack the 14 branches into eight leaf containers and assign each branch to exactly one container, preventing duplicated credit for shared prefixes. As in SPO-chain, tokens with generating-policy probability at least $0.9$ are masked from the loss. If tree collection fails, we discard the incomplete tree and sample a fresh group of eight standard GRPO trajectories.

\paragraph{CriticSearch.}
CriticSearch uses the same frozen \texttt{gpt-5.4-mini} as its retrospective critic. Because labeling every trajectory in every group would require many costly critic calls, we apply CriticSearch only to groups containing one, two, or three successes, matching \sys's setup; all other groups use GRPO. For each eligible group, the critic labels every trajectory, and CriticSearch uses
\begin{equation}
	A_t
	= 0.75 A_i^{\mathrm{GRPO}}
	+ 0.25 A_t^{\mathrm{critic}}.
	\label{eq:criticsearch}
\end{equation}
Because CriticSearch labels all eight trajectories in each eligible group, it uses eight critic calls per group, whereas \sys sends only one selected successful trajectory to its agentic judge. SearchQA critics receive reference answers and label valid search actions. ALFWorld and WebShop critics receive only policy-visible trajectories and terminal success, never hidden environment state. Invalid critic outputs or token alignment produce a whole-group GRPO fallback.

\stitle{Random segment selection.}
For the random-selection baseline introduced in Section~\ref{sec:agentic-judge-ablation}, we replace the agentic judge with random selection of a starting turn and a segment length between one and four turns from the selected successful trajectory.
Intervals that violate the proposal-validity constraints are discarded and receive no additional credit.
All other settings, including outcome verification and credit assignment, remain identical to \sys.

\subsection{Agentic Judge Prompt}
\label{app:agentic-judge-prompt}

The agentic judge receives the complete turn-indexed successful trajectory and compact indexes of the failed trajectories from the same GRPO group. The prompt asks it to diagnose a recurrent failure mode and return the shortest non-terminal expert segment that solves or avoids that failure. The judge is given trajectory search and bounded segment-inspection tools. Its output contains only the inclusive segment bounds and a rationale. We deterministically validate the output schema, turn bounds, action validity, and nonterminal endpoint, allowing up to three additional turns to correct any validation errors. 

\input{resources/agentic_judge_prompt}

%% file: resources/alfworld_prompt.tex
\noindent\begin{tcolorbox}[breakable, before skip=6pt, after skip=6pt, colback=white, colframe=black, title=ALFWorld Prompt Template, arc=3pt, boxrule=0.5pt, fonttitle=\normalfont, coltitle=white, top=2mm, bottom=2mm, left=2mm, right=2mm]
\begingroup
\csname ttfamily\endcsname\small\raggedright
\setlength{\parindent}{0pt}
\detokenize{SYSTEM}\par
\detokenize{You are controlling an agent in the ALFWorld household environment.}\par
\detokenize{The initial task, current room, and exact AVAILABLE ACTIONS are provided before your first turn.}\par
\detokenize{Call act with exactly one currently available action on every turn.}\par
\detokenize{Do not invent actions. Continue until the environment reports success or the turn budget ends.}\par
\medskip
\detokenize{USER}\par
\detokenize{Complete the ALFWorld task.}\par
\medskip
\detokenize{ENVIRONMENT OBSERVATION TEMPLATE}\par
\detokenize{OBSERVATION:}\par
\detokenize{{observation}}\par
\detokenize{AVAILABLE ACTIONS:}\par
\detokenize{- {action_1}}\par
\detokenize{- {action_2}}\par
\detokenize{...}\par
\detokenize{DONE: {done}}\par
\detokenize{REWARD: {reward}}\par
\endgroup
\end{tcolorbox}

%% file: resources/webshop_prompt.tex
\noindent\begin{tcolorbox}[breakable, before skip=6pt, after skip=6pt, colback=white, colframe=black, title=WebShop Prompt Template, arc=3pt, boxrule=0.5pt, fonttitle=\normalfont, coltitle=white, top=2mm, bottom=2mm, left=2mm, right=2mm]
\begingroup
\csname ttfamily\endcsname\small\raggedright
\setlength{\parindent}{0pt}
\detokenize{SYSTEM}\par
\detokenize{You are controlling an agent in the WebShop environment.}\par
\detokenize{The shopping instruction, current page, and available search/click actions are provided before your first turn.}\par
\detokenize{Call act exactly once on every turn. Use search[non-empty query] only when search is available, or click[target] with an exact current target.}\par
\detokenize{Continue until a purchase ends the episode or the turn budget ends.}\par
\medskip
\detokenize{USER}\par
\detokenize{Complete the WebShop task.}\par
\medskip
\detokenize{ENVIRONMENT OBSERVATION TEMPLATE}\par
\detokenize{OBSERVATION:}\par
\detokenize{{observation}}\par
\detokenize{AVAILABLE ACTIONS:}\par
\detokenize{- {visible_action_1}}\par
\detokenize{- {visible_action_2}}\par
\detokenize{...}\par
\detokenize{DONE: {done}}\par
\detokenize{SUCCESS: {success}}\par
\detokenize{SCORE: {native_reward}}\par
\endgroup
\end{tcolorbox}

%% file: resources/searchqa_prompt.tex
\noindent\begin{tcolorbox}[breakable, before skip=6pt, after skip=6pt, colback=white, colframe=black, title=SearchQA Prompt Template, arc=3pt, boxrule=0.5pt, fonttitle=\normalfont, coltitle=white, top=2mm, bottom=2mm, left=2mm, right=2mm]
\begingroup
\csname ttfamily\endcsname\small\raggedright
\setlength{\parindent}{0pt}
\detokenize{SYSTEM}\par
\detokenize{You are answering a question with the SearchQA environment.}\par
\detokenize{The exact question is provided before your first turn.}\par
\detokenize{Call act exactly once on every turn with exactly one argument named action.}\par
\detokenize{The action value must be one complete string with both tags: use <search>non-empty query</search> to retrieve evidence or <answer>concise final answer</answer> to finish.}\par
\detokenize{Never omit </search> or </answer>, and never send query or answer as a separate argument.}\par
\detokenize{Base the final answer on retrieved evidence and do not invent search results.}\par
\medskip
\detokenize{USER}\par
\detokenize{Answer the SearchQA question.}\par
\medskip
\detokenize{ENVIRONMENT OBSERVATION TEMPLATE}\par
\detokenize{OBSERVATION:}\par
\detokenize{{observation}}\par
\detokenize{AVAILABLE ACTIONS:}\par
\detokenize{- <search>non-empty query</search>}\par
\detokenize{- <answer>concise final answer</answer>}\par
\detokenize{DONE: {done}}\par
\detokenize{REWARD: {reward}}\par
\endgroup
\end{tcolorbox}

%% file: resources/agentic_judge_prompt.tex
\noindent\begin{tcolorbox}[breakable, before skip=6pt, after skip=6pt, colback=white, colframe=black, title=Agentic Judge Prompt Template, arc=3pt, boxrule=0.5pt, fonttitle=\normalfont, coltitle=white, top=2mm, bottom=2mm, left=2mm, right=2mm]
\begingroup
\csname ttfamily\endcsname\small\raggedright
\setlength{\parindent}{0pt}
\detokenize{SYSTEM}\par
\detokenize{You are an agentic judge identifying a short segment of a verified successful EXPERT trajectory that will receive additional fine-grained positive credit on top of the GRPO terminal reward.}\par
\medskip
\detokenize{Diagnose one recurrent failure mode from the failed trajectories, then select the shortest EXPERT segment that solves or avoids that diagnosed failure mode.}\par
\medskip
\detokenize{You receive the complete EXPERT trajectory directly in the user prompt and compact indexes for all failed trajectories from the same task. Each failed index gives its stable traj_id and inclusive start_turn/end_turn bounds for get_segment.}\par
\medskip
\detokenize{Diagnose a recurrent failure mode:}\par
\detokenize{The search_trajectory and get_segment tools are optional and available if useful. Identify a failure mode that recurs in multiple failed trajectories. Treat different actions as equivalent when they reach equally useful states, and do not group failures that require different fixes.}\par
\medskip
\detokenize{Select the EXPERT segment that solves the failure mode:}\par
\detokenize{From the complete EXPERT, find the shortest contiguous segment whose actions solve or avoid the diagnosed failure mode. A failure that already performs the proposed EXPERT transition is not valid contrastive evidence merely because it fails elsewhere.}\par
\medskip
\detokenize{The policy tokens in every selected turn will themselves receive additional positive credit. Select only actions whose own behavior should be reinforced. Judge each action together with the transition it causes, not merely by whether a later state is useful. A malformed, unavailable, unsuccessful, or error-producing action must not be selected merely because recovering from its error later helped the EXPERT succeed. A valid corrective action after an error may be selected when that corrective action itself is useful, but the preceding bad action must be excluded.}\par
\medskip
\detokenize{Segment-selection rules:}\par
\detokenize{1. Select the shortest contiguous segment that directly solves or avoids the diagnosed failure mode. Every selected turn must be necessary to that transition and worth reinforcing; exclude merely contextual or adjacent turns.}\par
\detokenize{2. Select behavior that materially advances the task, not behavior that is merely late in a successful trajectory. Do not select continued retrieval or repeated activity when the prior state is already sufficient unless it produces a necessary new state change.}\par
\detokenize{3. Treat actions as equivalent when they reach equally useful states, even if their routes, queries, object identities, or wording differ.}\par
\detokenize{4. The state after segment_end_turn must be nonterminal, and the terminal action must not be selected.}\par
\medskip
\detokenize{You do not write a hint, propose new actions, or assign reward. You only select an existing EXPERT segment. Use only recorded observations and affordances visible to the rollout policy. Deterministic validation separately checks the output schema, bounds, action grammar and affordances, recorded action outcomes, terminality, and sampled tokens; your responsibility is the semantic and causal judgment.}\par
\medskip
\detokenize{If you use search previews, treat them as bounded evidence rather than complete local behavior; use get_segment when more context would help. Failed trajectories are contrastive evidence, not demonstrations.}\par
\medskip
\detokenize{Return only one JSON object with exactly these fields:}\par
\char`\{\par
\detokenize{"segment_start_turn": <inclusive integer>,}\par
\detokenize{"segment_end_turn": <inclusive integer strictly before the EXPERT's final recorded turn, whose resulting state is nonterminal>,}\par
\detokenize{"rationale": "<the recurrent contrast, visible state before and after the selected segment, why each selected action itself deserves additional positive credit, and why this is the minimal complete causal transition>"}\par
\char`\}\par
\detokenize{Do not include markdown or any text outside the JSON object.}\par
\medskip
\detokenize{USER}\par
\char`\{\par
\detokenize{"task": {task},}\par
\detokenize{"expert": {complete_turn_indexed_expert_trajectory},}\par
\detokenize{"failed_trajectory_indexes": {compact_failed_trajectory_indexes},}\par
\detokenize{"instructions": "The expert is provided in full turn order above. Select one expert segment satisfying the system prompt. The trajectory tools are optional."}\par
\char`\}\par
\medskip
\detokenize{TOOLS}\par
\detokenize{search_trajectory(query, k=10): Globally rank bounded turn previews across all failed trajectories.}\par
\detokenize{get_segment(traj_id, start_turn, end_turn): Inspect a bounded inclusive turn interval from the expert or a failed trajectory.}\par
\endgroup
\end{tcolorbox}

%% file: sections/appendix/conditional_unbiasedness.tex
\section{Conditional Unbiasedness of the Segment Estimator}
\label{app:segment-estimator}

This section proves the conditional unbiasedness claim in Section~\ref{sec:verify} and characterizes the finite-sample variance of the raw pre/post estimator. The result does not apply to the positive part $[\widehat{\Delta}_{\mathrm{seg}}]_+$ or to the complete judge-selected training update.

\paragraph{Conditional unbiasedness.}
Let $\mathcal{F}$ contain the source rollout group and all information used to select $\tau^{\mathrm{seg}}$. Under exact boundary reconstruction, a fixed policy $\pi$, and conditionally independent continuation samples, the selected segment and its boundary states are fixed after conditioning on $\mathcal{F}$. Each empirical boundary value is an unbiased sample mean:
\begin{equation*}
    \mathbb{E}\!\left[\widehat{V}_{\mathrm{pre}}^{\pi}\mid\mathcal{F}\right]
    = V^{\pi}(s_{\mathrm{pre}}),
    \qquad
    \mathbb{E}\!\left[\widehat{V}_{\mathrm{post}}^{\pi}\mid\mathcal{F}\right]
    = V^{\pi}(s_{\mathrm{post}}).
\end{equation*}
Linearity of expectation and Equation~\ref{eq:segment-value-difference} therefore give
\begin{equation*}
\begin{aligned}
    \mathbb{E}\!\left[\widehat{\Delta}_{\mathrm{seg}}\mid\mathcal{F}\right]
    &= V^{\pi}(s_{\mathrm{post}})-V^{\pi}(s_{\mathrm{pre}}) \\
    &= A_{\mathrm{seg}}^{\pi}(s_{\mathrm{pre}},\mathbf{a}_{\ell:r}).
\end{aligned}
    \label{eq:delta-unbiasedness}
\end{equation*}
Thus, the raw $\widehat{\Delta}_{\mathrm{seg}}$ is conditionally unbiased for the selected segment's advantage. \hfill$\square$

\paragraph{Finite-sample variance.}
For binary terminal outcomes, let $p_{\mathrm{pre}}=V^{\pi}(s_{\mathrm{pre}})$ and $p_{\mathrm{post}}=V^{\pi}(s_{\mathrm{post}})$. The two sample means have conditional variances $p_{\mathrm{pre}}(1-p_{\mathrm{pre}})/K$ and $p_{\mathrm{post}}(1-p_{\mathrm{post}})/K$. Their conditional independence therefore gives
\begin{equation*}
    \operatorname{Var}(\widehat{\Delta}_{\mathrm{seg}}\mid\mathcal{F})
    = \frac{p_{\mathrm{post}}(1-p_{\mathrm{post}})
    + p_{\mathrm{pre}}(1-p_{\mathrm{pre}})}{K}.
    \label{eq:delta-variance}
\end{equation*}
Thus, the estimator becomes more precise as $K$ increases, with variance decreasing at rate $1/K$. For finite $K$, however, sampling noise can change the sign of $\widehat{\Delta}_{\mathrm{seg}}$, so a positive estimate provides empirical evidence of benefit rather than a conclusive guarantee.

\paragraph{Scope of the unbiasedness guarantee.}
The conditional-unbiasedness guarantee applies only to the raw estimator for a fixed segment after conditioning on the judge's selection. \sys instead uses its positive part,
\begin{equation*}
    [\widehat{\Delta}_{\mathrm{seg}}]_+
    = \max(\widehat{\Delta}_{\mathrm{seg}},0),
\end{equation*}
which is generally biased because negative sampling errors are replaced by zero while positive sampling errors are retained. Consequently,
\begin{equation*}
    \mathbb{E}\!\left[[\widehat{\Delta}_{\mathrm{seg}}]_+\mid\mathcal{F}\right]
    \neq
    [A_{\mathrm{seg}}^{\pi}(s_{\mathrm{pre}},\mathbf{a}_{\ell:r})]_+
\end{equation*}
in general. For example, even when the true segment advantage is zero, finite-sample estimates can be positive or negative; positive filtering discards the negative estimates but retains the positive ones, yielding a positive expected credit bonus. The complete \sys update additionally includes judge-guided segment selection, positive filtering, assigning the resulting bonus only to tokens within the selected segment, scaling by $\lambda$, and adding the bonus to the GRPO trajectory advantage. Therefore, conditional unbiasedness of the raw Monte Carlo estimator does not imply that the complete \sys update is an unbiased estimator of the standard policy gradient. Nor does the boundary comparison identify separate causal effects for individual actions or tokens within the segment.